\documentclass[letterpaper, 10 pt, conference]{ieeeconf}  

\usepackage{amsmath}
\usepackage{amssymb}
\usepackage{graphicx}

\IEEEoverridecommandlockouts                              

\title{\LARGE \bf
A Neural Network Model to Classify Liver Cancer Patients Using Data Expansion and Compression
}

\author{Ashkan Zeinalzadeh$^{1}$, Tom Wenska, Gordon Okimoto 
\thanks{$^{1}$Ashkan Zeinalzadeh was with the Cancer Research Center, University of Hawaii at Manoa,
        HI, USA
        {\tt\small azeinalz@nd.edu}}%
}

\begin{document}

\maketitle
\thispagestyle{empty}
\pagestyle{empty}

\begin{abstract}

We develop a neural network model to classify liver cancer patients into high-risk and low-risk groups using genomic data. Our approach provides a novel technique to classify big data sets using neural network models. We preprocess the data before training the neural network models. We first expand the data using wavelet analysis. We then compress the wavelet coefficients by mapping them onto a new scaled orthonormal coordinate system. Then the data is used to train a neural network model that enables us to classify cancer patients into two different classes of high-risk and low-risk patients. We use the leave-one-out approach to build a neural network model. This neural network model enables us to classify a patient using genomic data as a high-risk or low-risk patient without any information about the survival time of the patient. The results from genomic data analysis are compared with survival time analysis. It is shown that the expansion and compression of data using wavelet analysis and singular value decomposition (SVD) is essential to train the neural network model.

\end{abstract}


\section{INTRODUCTION}
The goal of this study is to build a neural network model to classify patients into high-risk and low-risk patients based on genomic data. To build this model we use the genomic data of $390$ patients. This model enables us to determine the risk status for a new patient without any knowledge about the patient's survival time, although the results of classification using the neural network must be comparable with the classification result from survival-time analysis.

The liver cancer data are big data sets. Neural network models are not computationally effective for big data sets. The first challenge to apply the neural network to the genomic data is the size of the data. The genomic data include over $20000$ genes. We define genes as the parameters of a desired signal. The high number of parameters increases the calculation complexity for the classification of the data in the neural networks \cite{Mohsen1}-\cite{Mohsen4}. The authors have developed an algorithm to reduce the number of genes (parameters) to less than $40$ \cite{Zeinalzadeh}-\cite{JAMMIT}. We consider the data as a matrix, in which rows are the genes and columns are samples (patients).

Based on the survival-time analysis, less than $17$ percent of patients are low-risk and the rest are high-risk. This type of data is called imbalanced data in the literature [2], in which one or a few subsets of the clustered data has significantly smaller size in comparison to the rest of subsets in the clustered data. There are different techniques to analyze the imbalanced data, e.g. regenerating the data by resampling \cite{Longadge}. The techniques for resampling from data are data dependent. Authors in \cite{Hadke}, develop a method to resample from data based on the statistics of the data. We do not regenerate the data using resampling. Building a neural network model to cluster imbalanced data is difficult, because the subsets of data with small sizes are ignored in the neural model. Despite the fact that the low-risk patients are a small subset of the patients, it is important not to recognize a low-risk patient as a high-risk patient because of the consequences of the subsequent treatment.

The data is composed of two different sets of censored data and uncensored data. The censored data corresponds to the patients for which survival time is unknown (live patients) and uncensored data corresponds to those for which survival time is known (deceased patients).

We compute the one-dimensional continuous wavelet coefficients for each patient separately. We perform the wavelet analysis using the Mexican hat wavelet. We vectorize the output of the wavelet and construct a new matrix by replacing each column of the original data matrix with the vectorized form of the wavelet coefficients. We map the data onto orthonormal bases, which are the left singular vectors of the new data matrix. The new data matrix is used to train a neural network model. The parameters of the neural network models are trained based on an iterative method. These parameters are optimized to provide classification results comparable with those from survival time analysis. It is shown that the expansion and compression of data leads to a better neural network model to classify the liver cancer patients.

The rest of the paper is structured as follows. In Section~\ref{sec:sta}, the survival time analysis is used to classify the patients. In Section~\ref{sec:wa}, the wavelet analysis and singular value decomposition is applied to expand the data to time-frequency waveforms and compress them respectively. In Section~\ref{sec:nnm}, a neural network model is trained to classify the patients. We describe how the parameters of the neural network model are optimized. In Section~\ref{sec:ROC}, we show the numerical results obtained through simulation evaluations. Finally, concluding remarks are provided in Section~\ref{sec:con}.

\section{METHODS}
\subsection{Survival-Time Analysis}\label{sec:sta}
\begin{figure}[h]

\centering
\framebox{\parbox{3.2in}{
\includegraphics[width=3in]{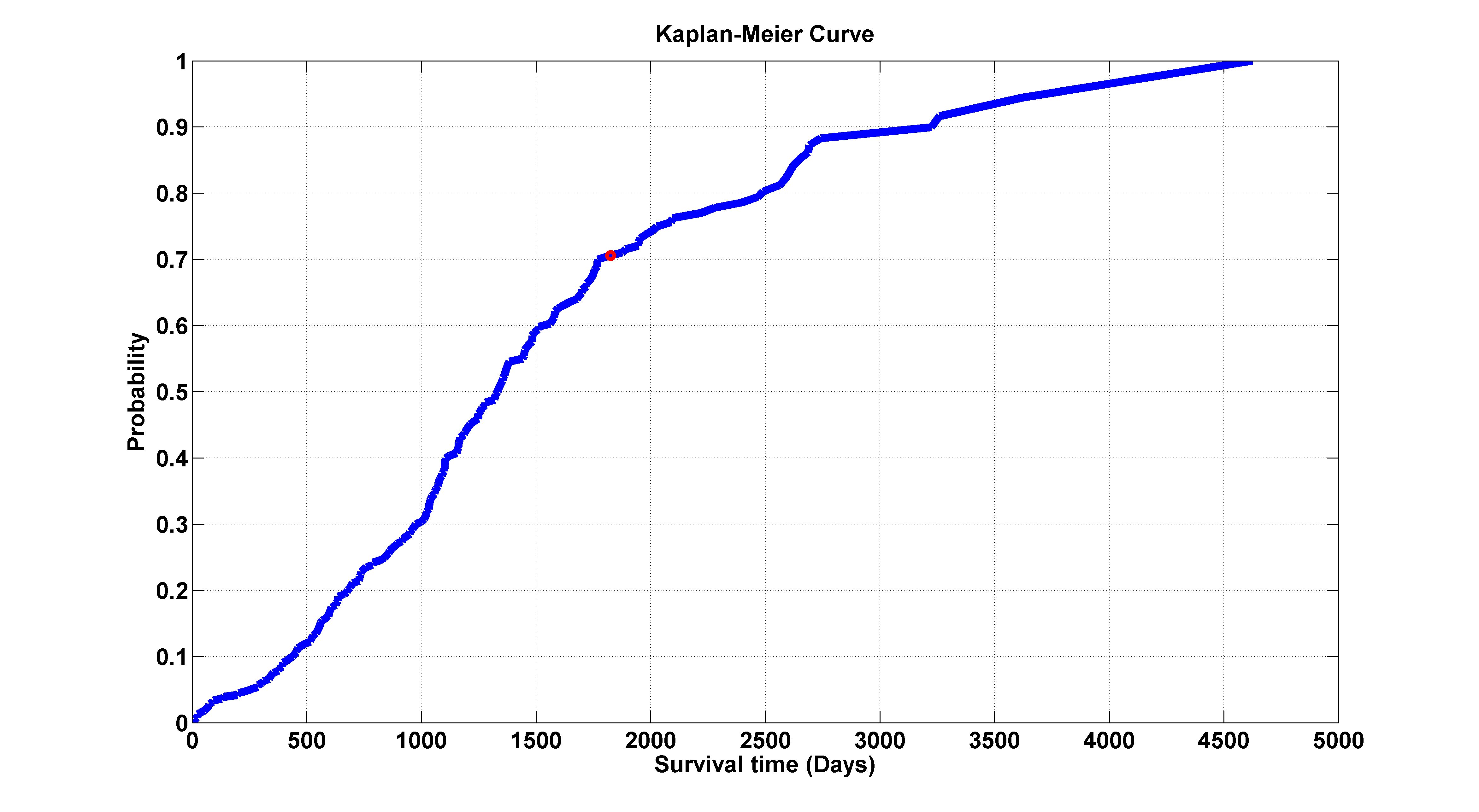}}}
\caption{The Kaplan-Meier CDF for the survival time of $390$ patients.}
\label{fig:kp}
\end{figure}

Patients with survival-time of more than $5$ years are called low-risk patients. The deceased patients (uncensored patients) with survival time less than 5 years are called high-risk patients. The risk status of a live patient (censored patient) with survival time, $ST$, of less than $5$ years is determined based on the Kaplan-Meier estimator, \cite{kap}. The advantage of the Kaplan-Meier estimator is to take into account the censored data. From $390$ patients, $173$ patients ($44\%$) are censored and $217$ patients ($56\%$) are uncensored. Let $ST$ be a random variable that denotes the survival time, $P$ be the Kaplan-Meier cumulative distribution function for $ST$. For a given $t$ less than $5$, the conditional cumulative distribution function (CDF) is defined as follows
\begin{align}\label{eq:one}
P(ST \geq 5|ST \geq t)=\frac{1-P(ST \leq 5)}{1-P(ST \leq t)}.
\end{align}
If the conditional CDF $P(ST \geq 5|ST \geq t)$ is equal to or bigger than $0.75$, then the patient is considered low-risk, otherwise the patient is considered high-risk. Out of $390$ patients, $67$ patients ($17\%$) are low-risk and $323$ ($83\%$) are high-risk. It is observed that a small number of patients are low risk patients. This can complicate the algorithm for training the neural network model to classify the patients. In figure~\ref{fig:kp}, the Kaplan-Meier CDF for $390$ patients is plotted. The vertical axis is the probability of survival time and the horizontal axis is the survival time in number of days. The red point in the figure~\ref{fig:kp} represents the five year survival time. It is observed that $P(ST \leq 5)=0.71$.

\subsection{Eigen Wavelet Features}\label{sec:wa}
Let $X=[x_1 \,\ x_2 \,\ \cdots \,\ x_n]$ be an $m \times n$ matrix. Each column of $X$ contains the data for one patient. The rows of the matrix $X$ correspond to genes. We use a continuous wavelet transform to analyze how the frequency content of a signal changes over a patient's genes. The wavelet transform compares the signal $x$ with shifted and scaled copies of a basic wavelet. We use the Mexican hat wavelet as the mother wavelet. This wavelet is very advantageous for analyzing genetic signals because of its explicit mathematical expression, smoothness, symmetry, and rapid decay. Let $\Psi(k)$ denotes the Mexican hat function of width $\sigma$
\begin{align}\label{eq:one}
\Psi(k)=\frac{2}{\sqrt{3\sigma}\pi^{\frac{1}{4}}} (1-\frac{k^2}{\sigma^2}) \exp^{\frac{-k^2}{2\sigma^2}}.
\end{align}
The larger the value of $\sigma$ the more the energy of $\Psi(k)$ is spread out over the genes (horizontal axis). The continuous wavelet transform (CWT) of a signal $x$ at a scale $s>0$ and position $\tau \in R$ is expressed by the following integral
\begin{align}\label{eq:two}
W_i(s,\tau;x_i(t),\Psi(t))=\frac{1}{\sqrt{|s|}} \int x(t) \Psi^{*} (\frac{t-\tau}{s}) dt,
\end{align}
where $*$ denotes the complex conjugate.
The CWT coefficients are affected by scale, $s$, position, $\tau$, and the mother wavelet function $\Psi$. In this analysis, the value of scale $s$ is fixed and the the position $\tau$ is between $1$ and $T$. An appropriate window size $T$ is chosen for the time-frequency localization. The value for $T$ is chosen by visually inspecting the wavelet coefficients. $W_i$ is a matrix of $T \times m$. Let $V_i$ be the vectorized form of the matrix $W_i$. We reconstruct the matrix $H=[ V_1 \,\ \cdots V_n]$. The wavelet coefficients $H$ is a matrix of $Tm \times n$. The wavelet transform ($\textit{w}$) expands the data from the space $m \times n$ to the space $Tm \times n$,
\begin{align}\label{eq:three}
X \overset{\textit{w}}{\longrightarrow}  H.
\end{align}
The expansion using wavelet analysis maps the data onto a larger space. It also expands a signal (through the genes) into a waveform whose time and frequency properties are the same as the original signal. We center the data matrix $H$ by taking the average of each row of data and subtracting it from each entry in that row.

We then compress the wavelet coefficients by mapping them onto a number of bases. These bases are the left singular vectors of the waveform coefficients $H$. The singular value decomposition of the matrix $H$ is given as
\begin{align}\label{eq:four}
H=LSR^T.
\end{align}
We project the data onto the first $k$ left singular vectors of the matrix $H$. Let $l_i$ and $r_i$ be the ith column of the orthonormal matrix $L$ and $R$ respectively
\begin{align}\label{eq:five}
\hat{H}=[l_1 \cdots l_k]^{T}H=[\sigma_1 r_1 \cdots \sigma_k r_k ]^T.
\end{align}
The mapping in (\ref{eq:five}), compresses the waveform coefficients onto a new scaled orthonormal coordinate system $[\sigma_1 r_1 \cdots \sigma_k r_k ]$,
\begin{align}\label{eq:six}
H \overset{[l_1 \cdots l_k]}{\longrightarrow}  \hat{H}.
\end{align}
Thus we shrink the size of the waveform coefficients from $Tm \times n$ to $k \times n$ and filter unwanted signals and noise. Choosing a right value for the parameter $k$ is crucial to classifying the patients correctly. The experimental results shows that the value of $k$ should be less than the number of genes $m$. We use an iterative method to choose a range of values for $k$ from $1$ to $m$. We finally select a value for $k$ that is able to classify the low-risk and high-risk patients efficiently. Choosing the value of $k$ depends on other parameters of the neural network model as will be described in the next section.

\subsection{Neural Network Model}\label{sec:nnm}
We first classify the patients into two groups of high risk and low risk patients using survival time analysis as explained in Section \ref{sec:sta}. We randomly choose $P$ percent of the high-risk and low-risk patients as the training data and $1-P$ percent of the high-risk and low-risk patients as the validation data. The value of $P$ is chosen based on an iterative method that will be explained in the next paragraph. The training and validation data are disjoint. The neural models are trained for $n$ separate times on all the patients' data except for one patient, and then prediction is made for that patient. The output of the neural network is a number between zero and one. We therefore choose a threshold between zero and one. If the output of the neural network is larger than the threshold, then the patient is considered high-risk, otherwise the patient is considered low-risk. Choosing the right value for the threshold is crucial for the analysis as described below.

The parameters for expansion, compression, and the training of the neural network are chosen based on an iterative method. These parameters are summarized as follows:
\begin{itemize}
\item Window size $T$ for the expansion of data using wavelet analysis.
\item Number of right singular vectors $k$ for the compression of data.
\item Percentage of the data $P$ that is used for the validation of the neural network model.
\item Number of hidden layers $\textit{h}$.
\item Threshold for the output of the neural network $Th$.
\end{itemize}
After iteration of the algorithm over a range of values, a set of values for these parameters are chosen to give us a high value of the true positive rate for low-risk patients and a small value for the false positive rate for the corresponding group of patients. The number of hidden layers $h$ must be smaller than or equal to the number of right singular vectors $k$. The threshold $Th$ is a number between zero and one.

\subsection{Validation Results}\label{sec:ROC}

\begin{figure}[h]

\centering
\framebox{\parbox{3.2in}{
\includegraphics[width=3in]{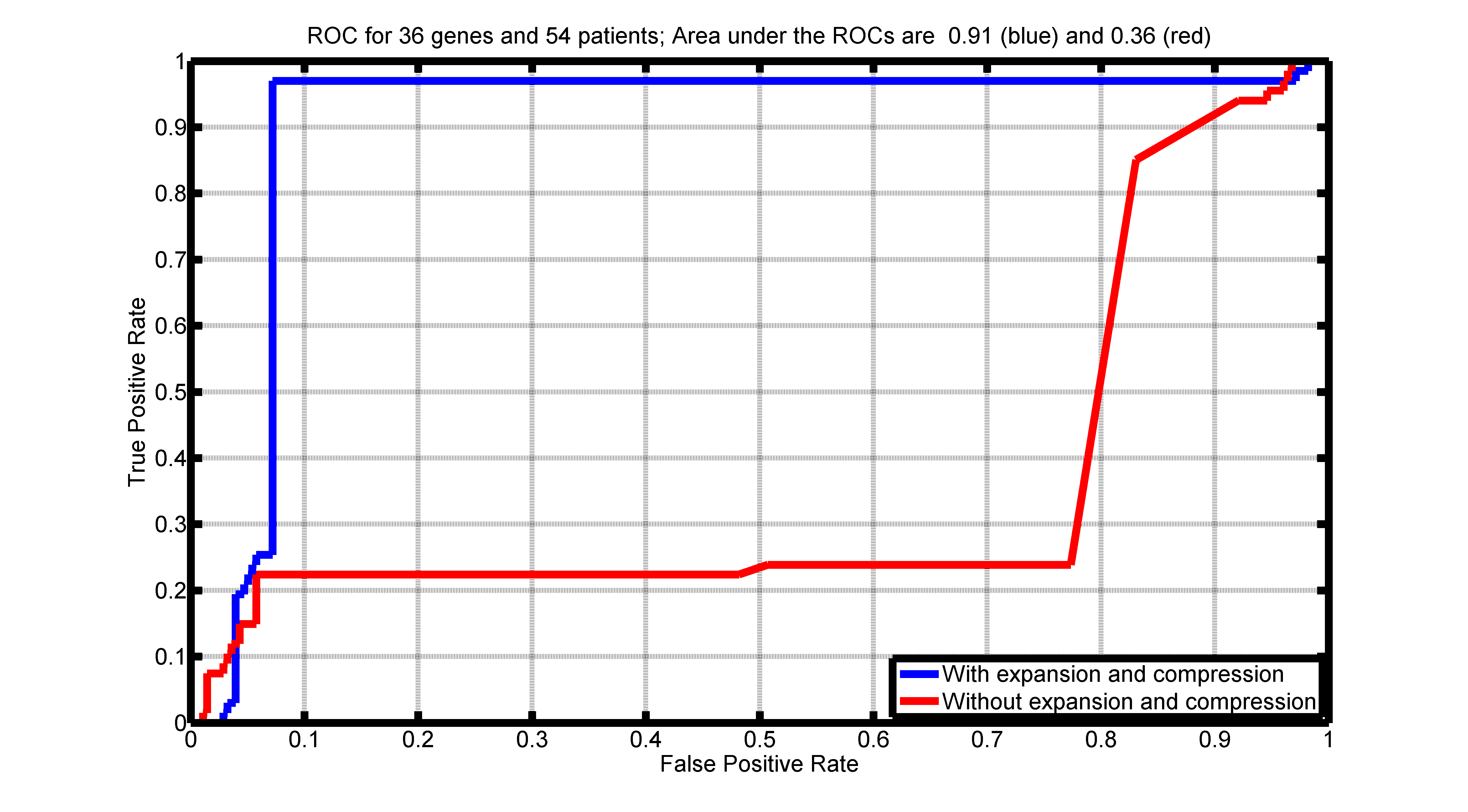}}}
\caption{ROC for $36$ genes and $54$ patients using leave-one-out approach.}
\label{fig:ROC1}
\end{figure}

\begin{figure}[h]

\centering
\framebox{\parbox{3.2in}{
\includegraphics[width=3in]{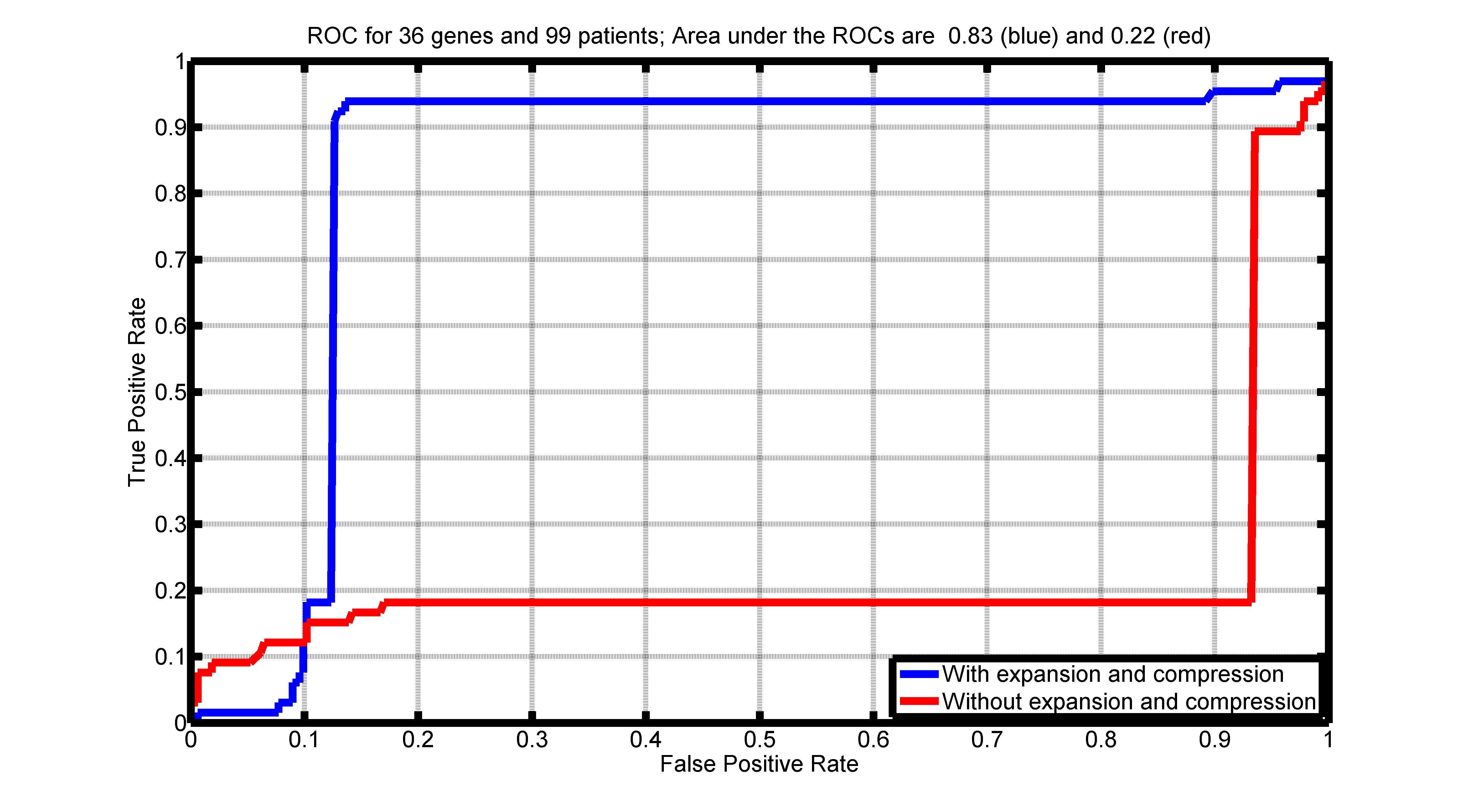}}}
\caption{ROC for $36$ genes and $99$ patients using leave-one-out approach.}
\label{fig:ROC2}
\end{figure}

\begin{figure}[h]

\centering
\framebox{\parbox{3.2in}{
\includegraphics[width=3in]{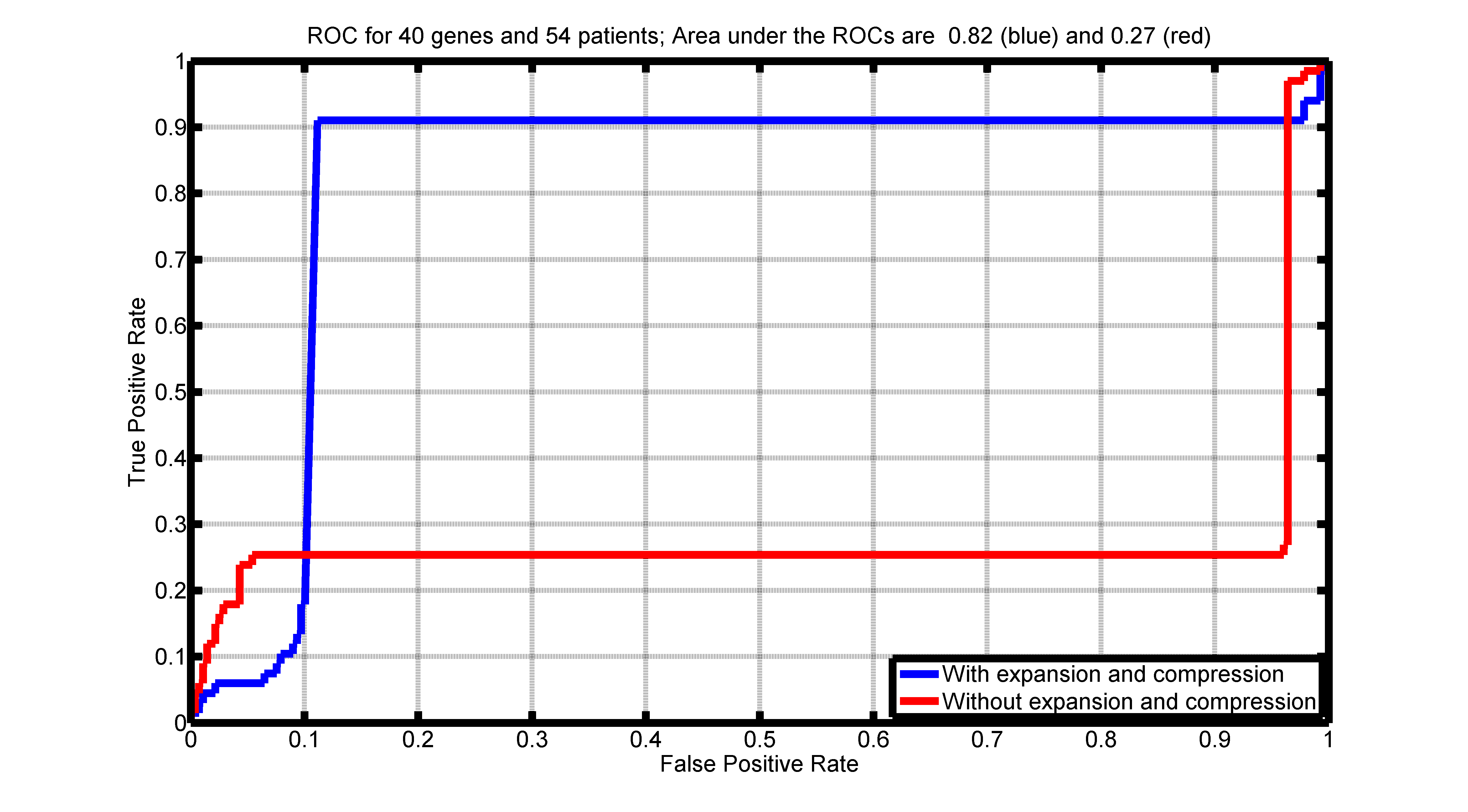}}}
\caption{ROC for $40$ genes and $54$ patients using leave-one-out approach.}
\label{fig:ROC3}
\end{figure}

\begin{figure}[h]

\centering
\framebox{\parbox{3.2in}{
\includegraphics[width=3in]{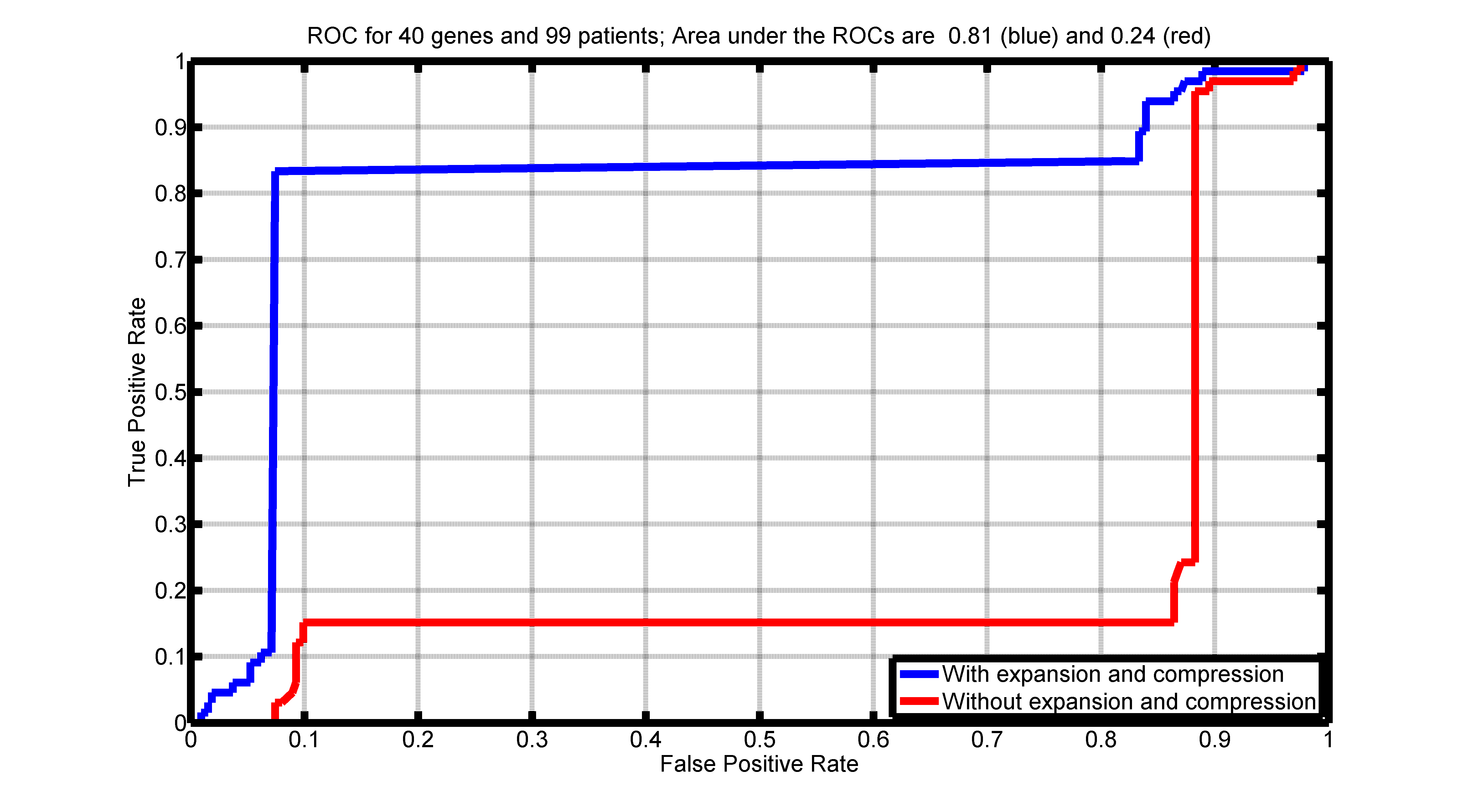}}}
\caption{ROC for $40$ genes and $99$ patients using leave-one-out approach.}
\label{fig:ROC4}
\end{figure}

In this work, false Positive Rate (FPR) is defined as the probability that a high-risk patient is recognized as a low-risk patient. True Positive Rate (TPR) is defined as the probability that a low-risk patient is recognized as a low-risk patient. Our class of interest is the low-risk patients. We consider two sets of genes, one including $36$ genes and another one including $40$ genes. These genes have been found by analyzing larger genomic data sets by the authors, as described in \cite{JAMMIT}. The genes in this signatures carry signal information that classify the ovarian cancer patients by their response to standard chemotherapy. We do the analysis for two separate groups of patients. The first group has $54$ patients. Out of $54$ patients, $20$ patients ($37\%$) are low-risk and the rest are high-risk.  The second group has $99$ patients. Out of $99$ patients, $19$ patients ($19\%$) are low-risk and the rest are high-risk. In figures~\ref{fig:ROC1}-\ref{fig:ROC4}, the receiver operating characteristics (ROCs) for the two sets of genes and two groups of patients are plotted. The vertical axis is the True Positive Rate and the horizontal axis is the False Positive Rate. An optimal threshold for the neural network model and a set of parameters for the data expansion and compression, is chosen based on an iterative algorithm as given in Table ~\ref{table:tb2}. The true positive rate, false positive rate, and area under the ROC, for the optimal parameters are given in Table~\ref{table:tb1}.

\begin{figure}[h]

\centering
\framebox{\parbox{3.2in}{
\includegraphics[width=3in]{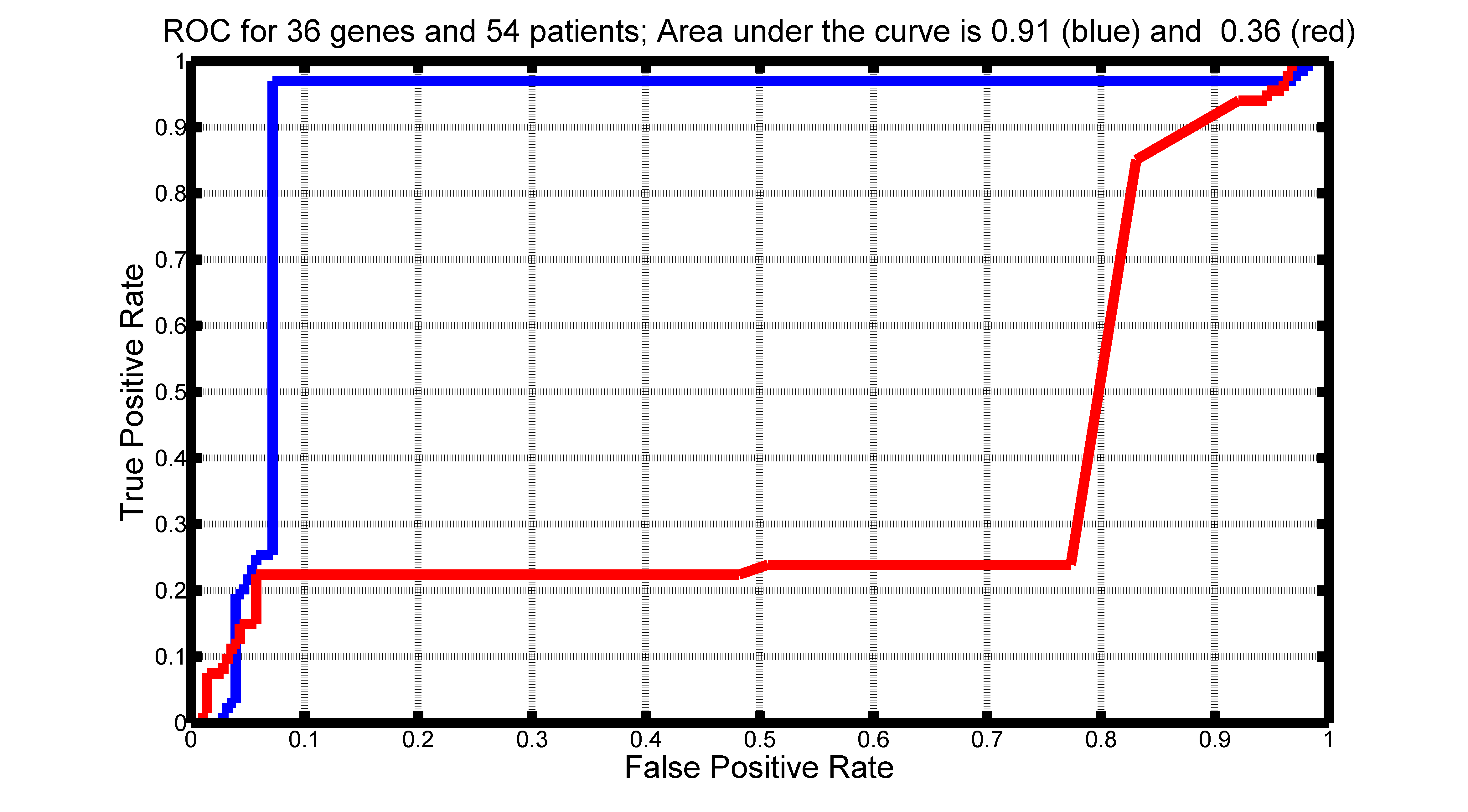}}}
\caption{ROC for $36$ genes and $54$ patients using regular neural network approach.}
\label{fig:ROC5}
\end{figure}

\begin{figure}[h]

\centering
\framebox{\parbox{3.2in}{
\includegraphics[width=3in]{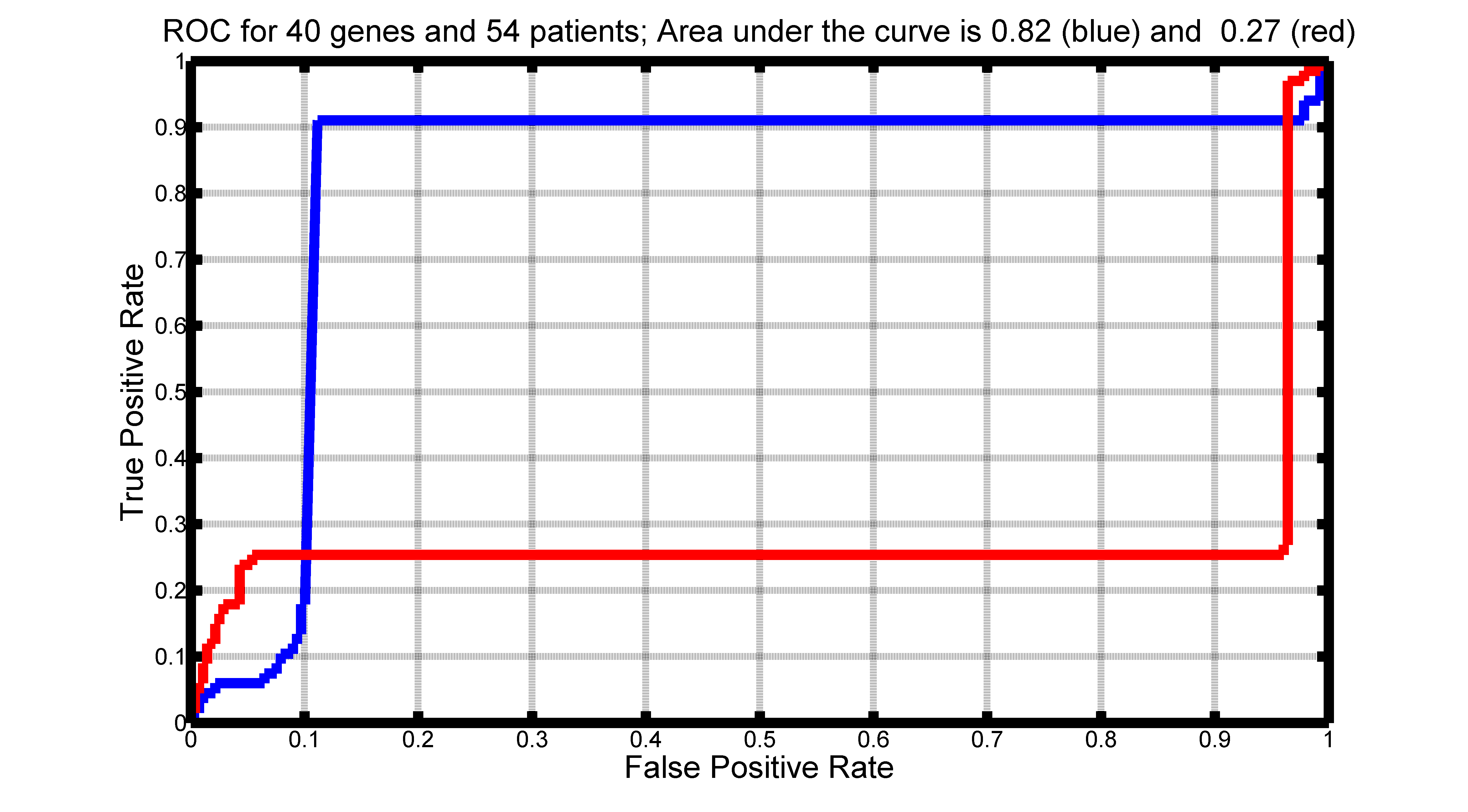}}}
\caption{ROC for $40$ genes and $54$ patients using regular neural network approach.}
\label{fig:ROC6}
\end{figure}

We second classify the patients into two groups of high risk and low risk patients using regular neural network model. We consider $390$ of patients as the training data. We train a model to classify $54$ patients. Out of $390$ patients, $67$ patients ($17\%$) are low-risk and $323$ ($83\%$) are high-risk. We randomly consider $164$ of high-risk patients and all the low-risk patients ($67$ patients) as the training data. The rest of high-risk patients are not considered in the analysis. We train the neural network model on $231$ patients, in which $29\%$ are low-risk and $71\%$ are high-risk. Then, prediction is made for $54$ patients. In figures~\ref{fig:ROC5}-\ref{fig:ROC6}, the receiver operating characteristics (ROCs) for the two sets of genes are plotted. It is observed that the result from leave-ne-out is similar to regular neural network when the number of high-risk patient is reduced.
Similar to the leave-one approach $P$ percent of the high-risk and low-risk patients as the training data and $1-P$ percent of the high-risk and low-risk patients as the validation data. The training and validation data are disjoint. Similarly, a threshold is chosen for the output of the neural network.

\begin{table}[h]
\caption{True Positive Rate (TPR), False Positive Rate (FPR), area under the ROC with expansion and compression (AREA1), and area under the ROC without expansion and compression (AREA2)}
\label{table:tb1}
\begin{center}
\begin{tabular}{|c|c|c|c|c|c|c|}
\hline
 Genes & Patients & TPR & FPR & AREA1 & AREA2\\
\hline
  $40$ & $99$   &   0.83 & 0.07 & 0.80 & 0.24\\
\hline
  $40$ & $54$   &   0.97 & 0.11 & 0.88 & 0.27\\
\hline
  $36$ & $99$   &   0.92 & 0.13 & 0.83 & 0.36\\
\hline
  $36$ & $54$   &   0.97 & 0.07 & 0.91 & 0.36\\
\hline
\end{tabular}
\end{center}
\end{table}


\begin{table}[h]
\caption{The parameters of expansion, compression and neural network}
\label{table:tb2}
\begin{center}
\begin{tabular}{|c|c|c|c|c|c|c|c|}
\hline
 Genes & Patients & T  & k   & P   & h & Th\\
\hline
  $40$ & $99$     & 5  & 7   &0.7  & 5 &0.84\\
\hline
  $40$ & $54$     & 5  & 7   &0.8  & 6 &0.83\\
\hline
  $36$ & $99$     & 5  & 7   &0.7  & 6 &0.84\\
\hline
  $36$ & $54$     & 5  & 7   &0.8  & 3 &0.83\\
\hline
\end{tabular}
\end{center}
\end{table}

\section{CONCLUSIONS}\label{sec:con}
It is observed that expansion and compression of the data, enable the neural network model to classify patients significantly better. The results from leave-one-out approach is comparable with results from regular neural network. Choosing the right parameters for the compression, expansion and training model is crucial for the analysis. Reducing the number of high-risk patients helps to train a neural network model to classify high-risk and low-risk patients.

\end{document}